\documentclass[10pt,twocolumn,letterpaper]{article}

\usepackage{cvpr}      

\usepackage{lipsum}  
\usepackage[dvipsnames]{xcolor}

\definecolor{cvprblue}{rgb}{0.21,0.49,0.74}
\usepackage[pagebackref,breaklinks,colorlinks,citecolor=cvprblue]{hyperref}

\usepackage{makecell}

\newcolumntype{B}{!{\hspace{-1ex}}c}
\newcolumntype{D}{!{\hspace{-2ex}}c}
\newcolumntype{A}{!{\hspace{-1ex}}l}
\usepackage{threeparttable}
\usepackage{multirow}
\usepackage{graphicx}
\usepackage{multicol}
\usepackage{comment}

\usepackage{amsmath}
\usepackage{amssymb}
\usepackage{cuted}
\usepackage{gensymb}
\usepackage{tablefootnote}

\def\paperID{***} 
\def\confName{CVPR}
\def\confYear{2024}

\title{TeTriRF: Temporal Tri-Plane Radiance Fields for Efficient Free-Viewpoint Video}

\author{
{Minye Wu 
\qquad 
Zehao Wang 
\qquad 
Georgios Kouros 
\qquad 
Tinne Tuytelaars} \\
{KU Leuven}  \\
}

\begin{document}
\maketitle

\begin{strip}\centering
\vspace{-17mm}
	\includegraphics[width=\linewidth]{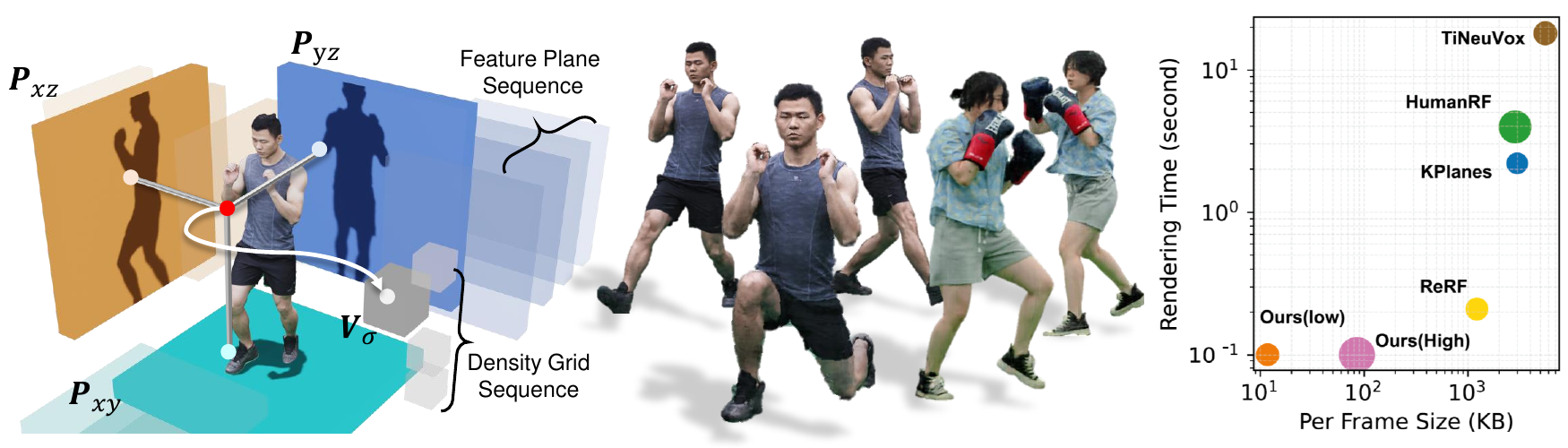}
	\vspace{-0.8cm}
	\captionof{figure}{\textbf{Left}: Temporal Tri-Plane Radiance Fields structure with plane-grid hybrid representation. \textbf{Middle}: TeTriRF results, showcasing novel view synthesis in dynamic scenes. \textbf{Right}: Comparison with existing methods in storage, speed, and quality, where larger circles indicate better quality. } 

	\label{fig:teaser}
	\vspace{-5mm}
\end{strip}

\begin{abstract}

Neural Radiance Fields (NeRF) revolutionize the realm of visual media by providing photorealistic Free-Viewpoint Video (FVV) experiences, offering viewers unparalleled immersion and interactivity. 
However, the technology's significant storage requirements and the computational complexity involved in generation and rendering currently limit its broader application. 
To close this gap, this paper presents Temporal Tri-Plane Radiance Fields (TeTriRF), a novel technology that significantly reduces the storage size for Free-Viewpoint Video (FVV) while maintaining low-cost generation and rendering. 
TeTriRF introduces a hybrid representation with tri-planes and voxel grids to support scaling up to long-duration sequences and scenes with complex motions or rapid changes. 
We propose a group training scheme tailored to achieving high training efficiency and yielding temporally consistent, low-entropy scene representations. 
Leveraging these properties of the representations, we introduce a compression pipeline with off-the-shelf video codecs, achieving an order of magnitude less storage size compared to the state-of-the-art. 
Our experiments demonstrate that TeTriRF can achieve competitive quality with a higher compression rate. 
Our project page is available at \href{https://wuminye.github.io/projects/TeTriRF/}{https://wuminye.github.io/projects/TeTriRF/}.

\end{abstract}    
\vspace{-4mm}
\section{Introduction}
\label{sec:intro}

Advanced VR/AR devices are boosting interest in Free-Viewpoint Video (FVV), which allows users to choose their own viewing angles for a unique and immersive exploration experience. 
The emergence of Neural Radiance Fields (NeRF), as introduced in~\cite{mildenhall2020nerf}, has significantly advanced FVV demonstrating unprecedented photorealism in rendering.
%
%
However, one main challenge with this technology, apart from its rendering speed, is the extensive storage space required for preserving reconstructed 4D data. This requirement complicates the process of transferring and storing such data on user devices, making the creation and use of long sequence FVV increasingly impractical.

Recent advances in NeRF facilitate dynamic scene rendering for FVV generation. 
Some models \cite{pumarola2021d,fang2022TiNeuVox,park2021nerfies,liu2022devrf} use deformation fields to model scene motion, mapping each frame to a canonical space. While these capture dynamics effectively, they are constrained by the high computational load of implicit feature decoding~\cite{pumarola2021d, fang2022TiNeuVox,park2021nerfies} or by the large storage needs of explicit 3D grid-based representations~\cite{liu2022devrf}. 
Alternatively, novel radiance field representations have been proposed to record dynamic scenes. They incorporate 4D data using techniques like planar factorization~\cite{fridovich2023k}, Fourier coefficients~\cite{wang2022fourier}, and latent embeddings~\cite{Li_2022_CVPR}. By training jointly across multiple frames, these methods achieve more efficient sequential frame reconstruction. 
However, their overly compact representation with limited capacity compromise their performance in capturing complex motions and long sequences. 
%
Most recently, several methods~\cite{song2023nerfplayer,peng2023representing,wang2023neural} have been developed to significantly improve the storage-performance trade-off. However, NeRFPlayer~\cite{song2023nerfplayer} suffers from a notably slow rendering speed, which prevents real-time playback for FVV. Dynamic MLP Maps~\cite{peng2023representing} and ReRF~\cite{wang2023neural}, on the other hand, achieve real-time rendering, but they require the use of a high-end GPU for decoding.

In this paper, we present a novel FVV modeling approach called Temporal Tri-Plane Radiance Fields or TeTriRF, which achieves efficient FVV generation and rendering with extremely compact storage. 
This is achieved via three main innovations. 
First, we propose a learning scheme that results in temporally consistent and low-entropy 4D sequential representations that can be effectively compressed. 
At the core is a training strategy that groups consecutive frames from sequential data and reduces the entropy of the frame representations via imposing  temporal consistency by deploying intra-group and inter-group regularizers. 
By sharing temporal information during training, TeTriRF is able to dramatically accelerate training compared to the per-frame training methods. 
We also deploy a two-pass progressive scaling scheme to reduce the cost of preprocessing while enhancing rendering quality and compression rate by discarding noise in empty space.

Second, we introduce a hybrid representation that combines tri-planes with voxel grids for frames within the sequence.
Specifically, for each frame in the stream, we factorize the radiance field to a tri-plane and a 3D density grid. 
This hybrid approach effectively captures high-dimensional appearance features in compact planes and enables efficient point sampling through the explicit density grid, achieving a balance between compactness and representation effectiveness. 
Building upon this hybrid representation, we adopt a deferred shading model~\cite{hedman2021baking, reiser2023merf} paired with lightweight MLP decoders to bring real-time rendering within reach.

Third, we show how to compress our FVV hybrid representation compactly using off-the-shelf video codecs. 
To achieve this, we develop a compression pipeline specifically for TeTriRF, which includes processes such as value quantization, removal of empty spaces, conversion into 2D serialization, and subsequent video encoding. 
The temporally consistent and low-entropy properties of our representation significantly enhance data compression efficiency. 
With our model, we're able to produce high-quality results with just 10-100 KB/frame. 
This means that a one-hour video could be stored in 1-10 GB, which is for the first time, within range of memory available on AR/VR devices. 
TeTriRF, with its compact size and our hybrid representation, is capable of handling long sequence FVV effectively. 
Together with our lightweight renderer and hardware accelerated video decoding, our approach takes another step towards streaming and rendering photorealistic FVV for end-users.


%

%

%
%
%
Fig.~\ref{fig:teaser} illustrates TeTriRF's representation structure and comparison with other methods. 
TeTriRF distinguishes itself with superior advantages in storage efficiency, rendering speed, and rendering quality.


%


\section{Related Work}
\label{sec:related_work}

\textbf{Neural Scene Representations.} 
NeRF~\cite{mildenhall2020nerf} achieves photo-realistic novel view synthesis using a simple implicit representation. 
Despite the quality and compactness of NeRF, the scene reconstruction and rendering times are substantial and prohibitive for the reconstruction of both static and dynamic scenes.
Subsequent works distill the volumetric representation of the scene into voxel grids to achieve real-time rendering speeds~\cite{liu2020neural,yu2021plenoctrees,hedman2021baking} or also fast volumetric scene reconstruction~\cite{yu_and_fridovichkeil2021plenoxels, sun2022dvgo}. 
Nonetheless, these approaches encounter the drawback of increased storage size due to the use of 3D grid representations. 
%
%
These issues have partly been mitigated by substituting the 3D voxel grid representation with more compact and memory-efficient tensor decompositions~\cite{chen2022tensorf}.
Further improvements in training and rendering speed have also been achieved by leveraging trainable multi-resolution hash tables \cite{mueller2022instant} or representations based on 3D Gaussians \cite{kerbl3Dgaussians}.
%
%
%
While efficient and compact photo-realistic reconstruction and rendering for static scenes are now achievable, the main challenge remains in dynamic scenes. In such scenarios, storage requirements typically increase linearly with the number of frames, leading to greater difficulties.

\textbf{Dynamic Radiance Field Representations.} 
Extending radiance fields to enable the reconstruction of dynamic scenes increases the requirements with regard to efficiency, compactness, and ability to handle long-duration sequences.
Some approaches reconstruct dynamic scenes by conditioning an implicit representation on time \cite{du2021nerflow,Gao-ICCV-DynNeRF,xian2021irrad} or time-varying latent codes \cite{Li2021Neural3V,peng2023representing}.
Alternatively, other methods \cite{pumarola2021d,park2021nerfies,du2021nerflow,Li2021Neural3V,Tretschk2020NonRigidNR,Zhang2021EditableFV,fang2022TiNeuVox,park2021hypernerf} optimize a deformation field to predict the displacement of the scene across time between each frame and a reference canonical frame.
%
%
%
%
%
Most implicit time-conditioned or deformation-based methods suffer from slow training and rendering speeds.
To accelerate the speeds, methods have been developed using grid representations \cite{Guo_2022_NDVG_ACCV,liu2022devrf}, 4D plane-based representation~\cite{xu20234k4d,shao2023tensor4d,cao2023hexplane}, and tensor factorization~\cite{fridovich2023k,Cao2022FWD,isik2023humanrf}. Even though they provide faster training or rendering times but they usually suffer from storage efficiency or network capacity issues.
%
%
%
%
%
Our proposed method resembles K-planes~\cite{fridovich2023k}, a method based on planar factorization. However, rather than using an additional tri-plane for spatio-temporal variations we simply unfold a triple-plane representation along the temporal dimension. Combined with our training and compression schemes, our method achieves improved rendering speed and compactness with competitive quality.

\textbf{Neural Radiance Field Compression.} 
Implicit neural representations like NeRF~\cite{mildenhall2020nerf} are relatively compact but at the same time extremely slow to train and render.
%
Improved compression in the context of neural radiance fields has been achieved via vector quantization~\cite{li2023compressing},  wavelet transforms on grid-based neural fields~\cite{rho2023maskedwavelet}, parameter pruning~\cite{deng2023compressing}, and Fourier transform~\cite{huang2022pref}. These approaches are primarily focused on static scenes and lack the capability to compress temporal information.
Recent works extend the compression into dynamic scene by using tensor decomposition~\cite{song2023nerfplayer} , residual radiance fields with specialized video codecs~\cite{wang2023rerf}, and reducing  spatio-temporal redundancies of feature grids \cite{guo2023compact}. 
However, fast decoding and rendering time is still an issue with these methods. Our method, on the other hand, achieves both high compression and fast decoding and rendering times thanks to our proposed representation, lightweight renderer, and off-the-shelf encoding.

\section{Methodology}

Our method is able to generate free-viewpoint videos from synchronized multi-view video inputs to support novel view synthesis in dynamic scenes with long-duration sequences and complex motions.
We represent each frame using a specialized hybrid representation with disentangled and compactly organized geometry and appearance, complemented by an efficient rendering pipeline.
%
We propose a fast training scheme (Sec.~\ref{sec:training}) that trains multiple frames in groups and lowers the entropy of features by improving their temporal consistency.
This facilitates extremely compact compression in our proposed pipeline. We demonstrate that these representations can be efficiently compressed using off-the-shelf video codecs (Sec~\ref{sec:compression}).


\begin{figure}[t]
	\begin{center}
		\includegraphics[width=1.0\linewidth]{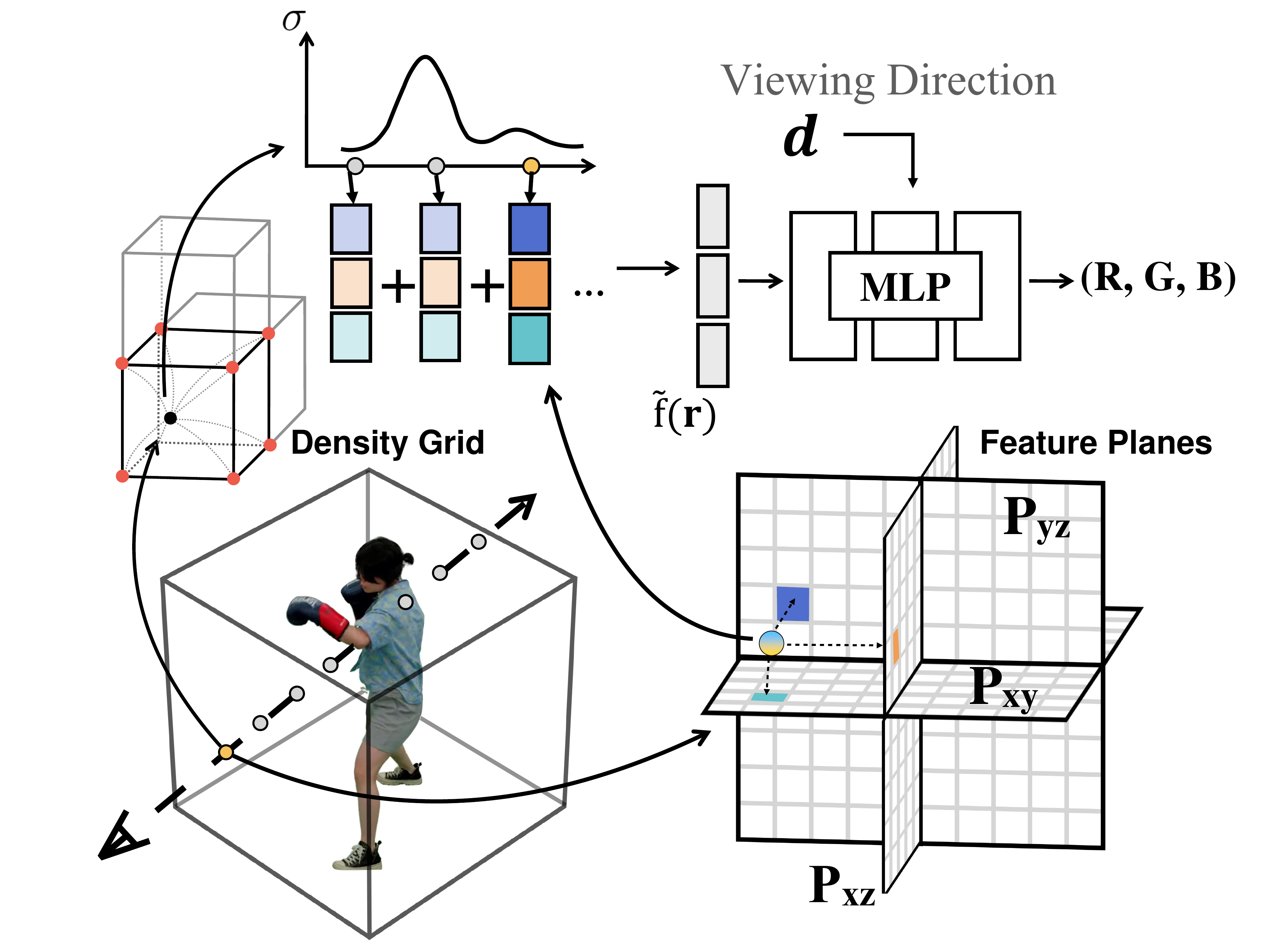} 
	\end{center}
  \vspace{-6mm}
	\caption{\textbf{Illustration of the Hybrid Representation.} Our Hybrid Tri-Plane approach models each frame using a density grid and a tri-plane. We adopt the deferred shading model in our rendering pipeline. 
 }  
	\label{fig: 31}
 \vspace{-4mm}
\end{figure}

\begin{figure*}[t]
	\begin{center}
        \includegraphics[width=1\linewidth]{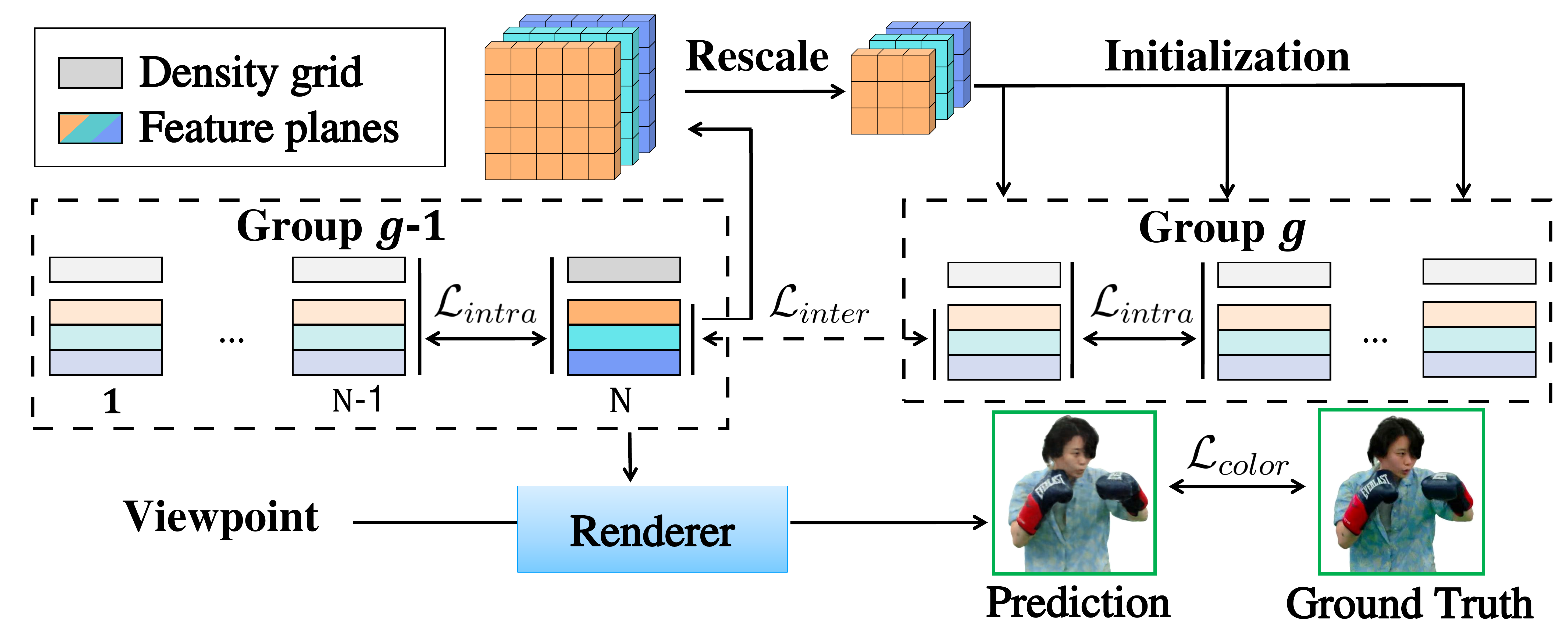} 
	\end{center}
  \vspace{-6mm}
	\caption{\textbf{Grouped Multi-frame Training Overview.} TeTriRF group frames in sequential data and trains a group of $N$ frames together. We deploy photometric loss $\mathcal{L}_{color}$, intra-group loss $\mathcal{L}_{intra}$, and inter-group loss $\mathcal{L}_{inter}$ among the frames. For a given viewpoint, the Renderer (Sec. \ref{sec:tri_plane}) takes the corresponding frame's representation to synthesize novel views.  
 }
	\label{fig: 32}
  \vspace{-4mm}
\end{figure*}

\subsection{Hybrid Tri-Plane}\label{sec:tri_plane}

In this work, as illustrated in Fig.~\ref{fig: 31}, we use a hybrid representation composed of a 3D density grid $\mathbf{V}_{\sigma}$ and a feature Tri-Plane $\mathbf{P}=\{\mathbf{P}_{s} | s \in \mathcal{S} \}$ to represent each frame, where each element in $\mathbf{P}$ is a 2D grid with $h=10$ channels and $\mathcal{S} = \{xy,xz, yz\}$. 
The purpose of this hybrid design is to attain a good trade-off between effectiveness and compactness.
The utilization of an explicit density grid allows the direct and fast acquisition of density values.
This enables the straightforward construction of a grid mask to efficiently discard sample points in free space without network inference and hence speed up both training and rendering. 
%
%
A feature Tri-Plane, on the other hand, contains three orthogonal feature planes factorizing the spatial space of higher-dimensional appearance features.
We adopt the compactness of this plane-based representation, which has been shown in \cite{fridovich2023k}, to elevate the compression rate to a new level.

%

\textbf{Rendering.} 
TeTriRF queries the density $\sigma$ of a 3D point $\mathbf{x}$ by applying trilinear interpolation $\varphi_t(\cdot)$ on the 3D density grid $V_{\sigma}$. 
Appearance features, on the other hand, are acquired by projecting the point onto the three feature planes $\mathbf{P}$ and applying bilinear interpolation $\varphi_b(\cdot)$ for each 2D projection. These operations are formulated as: 
\begin{equation}
\begin{split}
\sigma &= \varphi_t(\mathbf{x}, V_{\sigma}) \\
\mathbf{f}_s &=  \varphi_b(\mathbf{x}, \pi_s(\mathbf{P}_s)),
\end{split}
\label{eq:fetch}
\end{equation}
where $\pi_s$ is a function that projects a 3D point onto plane $s$, and $\mathbf{f}_s$ is the fetched $h$-dimensional feature vector from that plane. 
Then we concatenate the appearance features from the three planes to compose the feature vector $\mathbf{f}= [\mathbf{f}_{s} | s\in\mathcal{S} ]$. 
%
%

%
%
Since having the explicit density grid, we adopt the masking mechanism from DVGO to discard points in empty space and thus reduce the associated overhead in acquisition, processing, and volume rendering. 
Inspired by \cite{hedman2021baking, reiser2023merf}, we also adopt the deferred shading model which performs volume rendering on the appearance features of sample points along a ray, rather than their radiance. 
%
%
The radiance value (pixel RGB color) $\mathbf{c}$ of the ray is then decoded through a shallow MLP:
\begin{equation}
\mathbf{c(r)} = \Phi(\mathbf{\tilde{f}(r)}, \omega(\mathbf{d})), 
\end{equation}
where $\Phi$ denotes the decoding operation performed by the shallow MLP, responsible for transforming appearance features into radiance and $\tilde{\mathbf{f}}(\mathbf{r})$ is the integrated feature vector form the deferred shading model.
Additionally, $\mathbf{d}$ denotes the viewing direction of the ray and $\omega(\cdot)$ represents the positional encoding used in~\cite{mildenhall2020nerf}. 
This deferred rendering approach is also differentiable as gradients are backpropageted from the RGB color to the tri-plane, density grid, and MLP network parameters.


\subsection{Grouped Multi-frame Training}\label{sec:training}

Acquiring hybrid Tri-Plane representations for sequential frames is a non-trivial task. Per-frame training, as employed in ReRF~\cite{wang2023neural}, lacks efficiency. This is mainly because ReRF only exploits information from adjacent frames
In our method, we train a group of $N$ consecutive frames jointly at one time to more effectively leverage temporal information.
We train sequential data in non-overlapping groups according to the timeline, enabling TeTriRF to support long or even infinite sequences.
To promote information sharing and reduce redundancy during optimization we regularize the groups at both the intra- and inter-level as detailed in the following paragraphs.

\textbf{Intra-Group Regularization.} 
We apply L1 loss on density grids and feature planes between adjacent frames to encourage sparsity and minimize density and feature changes. This is crucial because video codecs are tasked with encoding the differences between frames. Consequently, sparsifying and minimizing these changes, effectively reduces the bitrate required for video encoding, leading to more efficient data compression. 
Temporal information, such as cross-frame density variations, can also be passed and shared among frames resulting in faster training.
We formulate this intra-group regularization as:
\begin{equation}
\begin{split}
\mathcal{L}_{intra} = \sum_{i=1}^{N-1} \Biggl[ \left\|V_\sigma^{g,i} - V_\sigma^{g,i+1}\right\|_1  
+  \sum_{s \in \mathcal{S}} \left\| P_{s}^{g,i} - P_{s}^{g,i+1} \right\|_1  \Biggr],
\end{split}
\end{equation}
where $g$ and $i$ denote the group index of these frames and the frame index inside the group, respectively. 
%
Furthermore, sharing the MLP decoder within the group defines a shared appearance feature space that facilitates faster convergence.

\textbf{Inter-Group Regularization.} 
To reduce calculation redundancy and speed up optimization, we initialize every frame in the current group with the feature planes from the last frame of the previous group as shown in Fig.~\ref{fig: 32}. This approach leverages existing information to provide an advantageous starting point for the training process. The MLP decoder is also initialized with the parameters of the decoder from the previous group for the same reason.

In addition to the initialization, we also apply an L1 feature loss between the first frame of the current group and the last frame of the previous group, ensuring feature continuity between the groups and thus increasing the compression rate. This L1 cross-group feature loss is formulated as:
\begin{equation}
\mathcal{L}_{inter} = \left\|V_\sigma^{g-1,N} - V_\sigma^{g,1}\right\|_1  
+  \sum_{s \in \mathcal{S}} \left\| P_{s}^{g-1,N} - P_{s}^{g,1} \right\|_1.
\end{equation}
We block the gradients to $V_\sigma^{g-1,N}$ and $P_{s}^{g-1,N}$ from the previous group to keep their representation consistent after being trained. 
This allows to train one group of frames at a time.

Along with the photometric loss $\mathcal{L}_{color}$, the total loss $\mathcal{L}_{total}$ is calculated as: 
\begin{equation}
\mathcal{L}_{total} = \mathcal{L}_{color} + \lambda_1\mathcal{L}_{intra} + \lambda_2\mathcal{L}_{inter}, 
\end{equation}
where $\lambda_1$ and $\lambda_2$ are the weights of the regularizing terms.

\textbf{Two-pass Progressive Scaling.} 
DVGO \cite{sun2022dvgo} and ReRF \cite{wang2023rerf} follow a two-stage coarse-to-fine training scheme. In the first stage, diffuse color is used to reconstruct a coarse density field for building a grid mask used to discard empty space. This strategy helps prevent the occurrence of numerous floaters and enhances the training speed.
Likewise, HumanRF \cite{isik2023humanrf} precomputes an occupancy grid for making training more efficient. However, the occupancy calculation highly depends on the camera setups, requiring the number of visible views for each space to be evenly distributed.
%
All these extra procedures take extra time and resources to process.

In our approach, we also maintain occupancy grids as in 
\cite{wang2023rerf,isik2023humanrf}, but we make them more adaptive and efficient for dynamic scenes regardless of the camera setups.
To this end, we introduce a two-pass progressive scaling strategy, where at predefined iterations we rescale the resolution of the density grid and feature planes as in~\cite{sun2022dvgo}. The first pass upscales the space at shorter time intervals, functioning similarly to the coarse training stage in scene space exploration. 
Upon completing the first pass, we revert the scale to its initial, lowest resolution. This reduction in resolution diminishes the presence of floaters, thereby increasing the availability of empty space. Most importantly, as we maintain the same MLP decoder throughout the training process, the down-scaled feature planes from the first pass can be effectively reused. This serves as an effective initialization for subsequent training phases, diverging from previous coarse-to-fine training strategies where appearance features are typically retrained from scratch. 
Following the second progressive scaling pass, we update the training rays by filtering out those that do not hit occupied space inside the bounding box, thus focusing the training on the reconstruction of fine detail. 
These training strategies can efficiently remove empty space regardless of camera setups and improve results with the same number of iterations as the fine stage of ReRF even without a coarse training stage. 
Please refer to our supplementary materials for more details.

\begin{figure}[t]
	\begin{center}
		\includegraphics[width=1.0\linewidth]{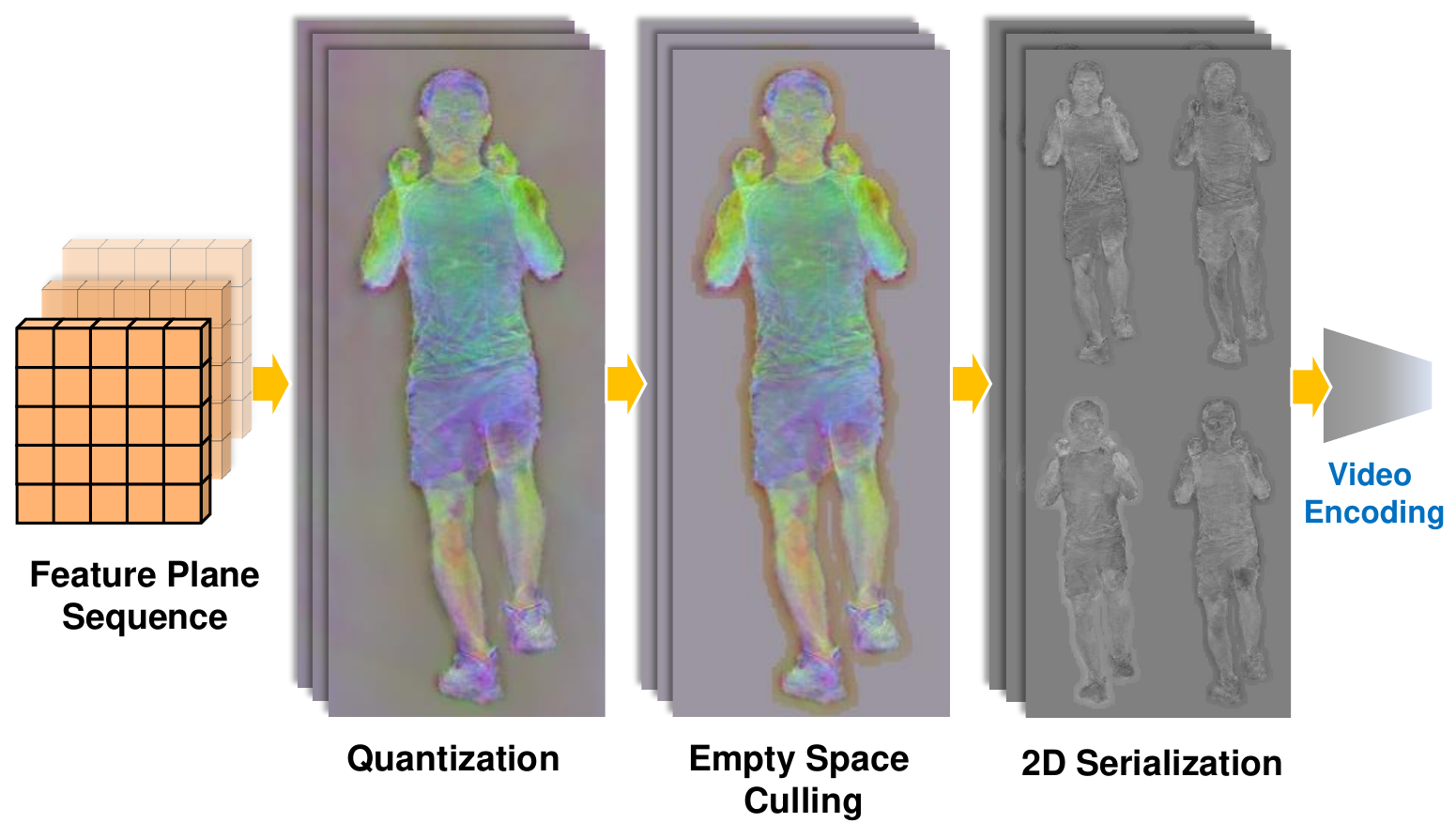} 
	\end{center}
  \vspace{-5mm}
	\caption{ Demonstration of the proposed compression pipeline and the visualizations after each process on a feature plane sequence. Colorized by the first three principal components. }  
	\label{fig:compression}
 \vspace{-4mm}
\end{figure}

\begin{table*}[ht]

\begin{tabular}{B|BBBBBB|BBBBBB} \hline 
              & \multicolumn{6}{c |}{NHR Dataset}                                                                        & \multicolumn{6}{c}{ReRF Dataset}                          \\ \midrule
              & PSNR $\uparrow$  & SSIM $\uparrow$  & LPIPS $\downarrow$ & \makecell{T.T. \\ (min)} $\downarrow$ & \makecell{R.T. \\ (s)} $\downarrow$ & \makecell{Size \\ (KB)} $\downarrow$ & PSNR $\uparrow$  & SSIM $\uparrow$  & LPIPS $\downarrow$ & \makecell{T.T. \\ (min)} $\downarrow$ & \makecell{R.T. \\ (s)} $\downarrow$ & \makecell{Size \\ (KB)} $\downarrow$ \\ \midrule
KPlanes & 30.18 & 0.963 & 0.063 & 0.65                       & 2.2                      & 2986                    & 27.81 & 0.946 & 0.094 & 0.65                       & 2.2                      & 2986                    \\
HumanRF & 31.91 & 0.872 & \textbf{0.036} & 1.4                        & 3.95                     & 2852                    & 28.58 & 0.876 & 0.072 & 1.4                        & 3.95                     & 2852                    \\

TiNeuVox      & 30.45 & 0.962 & 0.077 & 2.4                        & 18.14                    & 5580                    & 28.86 & 0.947 & 0.082 & 2.8                        & 23.65                    & 5580                    \\
ReRF          & 30.34 & 0.972 & 0.055 & 21.2                       & 0.21                     & 1220                    & \textbf{30.33} & \textbf{0.962} & \textbf{0.054} & 22.4                       & 0.27                     & 843                     \\ \midrule
Ours (low)    & 30.42 & 0.966 & 0.059 & \multirow{2}{*}{\textbf{0.55}}      & \multirow{2}{*}{\textbf{0.1}}    & \textbf{11.76}                   & 27.60 & 0.950 & 0.083 & \multirow{2}{*}{\textbf{0.58}}      & \multirow{2}{*}{\textbf{0.12}}    & \textbf{11.72}                   \\
Ours (high)   & \textbf{32.57} & \textbf{0.978} & 0.045 &                            &                          & 85.33                   & 30.18 & \textbf{0.962} & 0.056 &                            &                          & 71.67  \\ \hline    \hline              
\end{tabular}
\vspace{-2mm}
\caption{Results on NHR~\cite{wu2020multi} and ReRF~\cite{wang2023rerf} datasets. Training time (T.T.), Rendering Time (R.T.), and sizes are averaged over frames.}
\label{tab: 1}
\vspace{-2mm}
\end{table*}

\begin{figure*}[t]
	\begin{center}
		\includegraphics[width=1.0\linewidth]{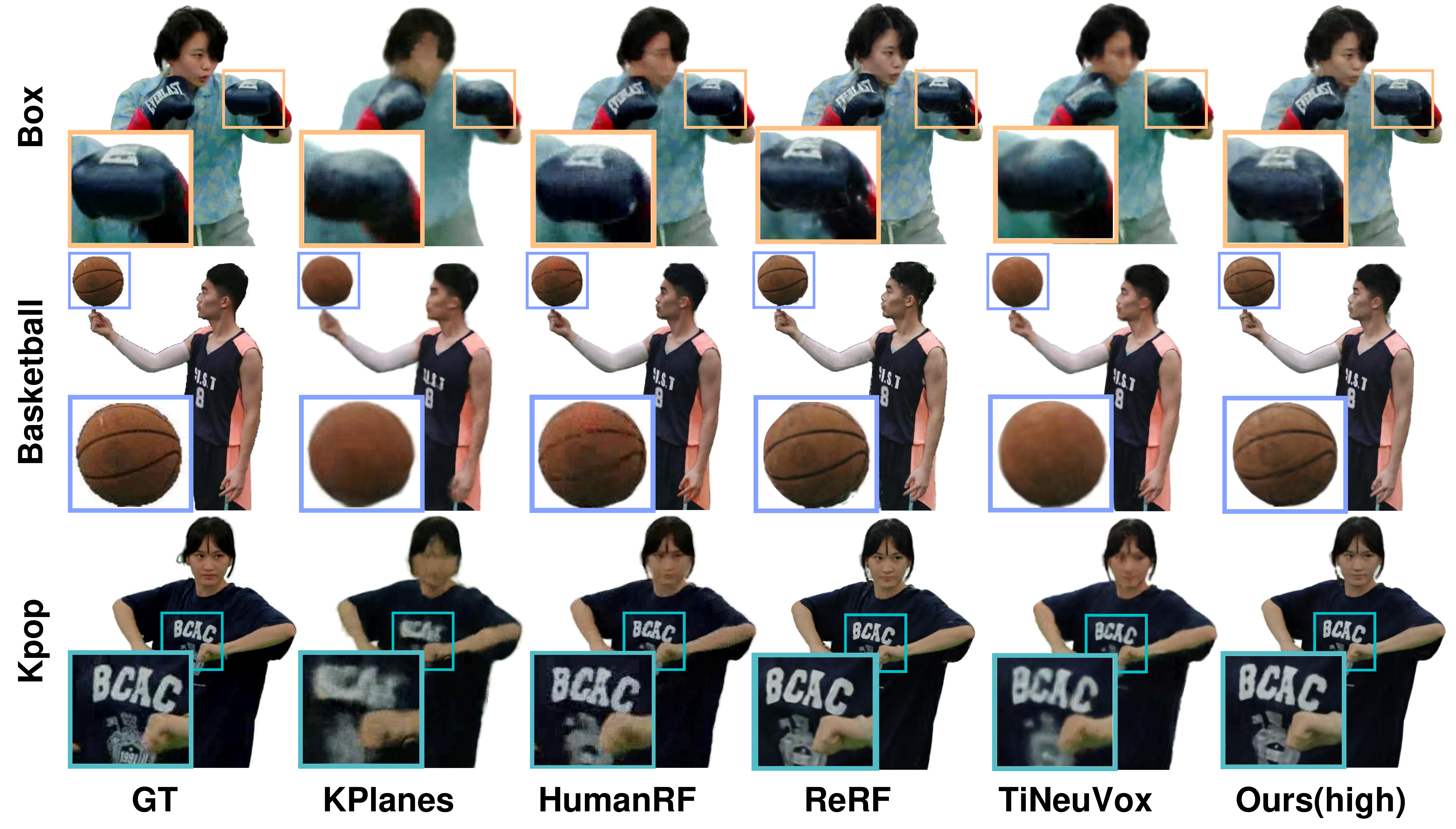} 
	\end{center}
 \vspace{-5mm}
	\caption{Qualitative comparison. The rendering quality of TeTriRF compared to four baselines KPlanes~\cite{fridovich2023k}, HumanRF~\cite{isik2023humanrf}, ReRF~\cite{wang2023rerf}, TiNeuVox~\cite{fang2022TiNeuVox} on human-centric scenes \textit{box}, \textit{basketball} and \textit{kpop} from NHR and ReRF datasets. See supplemental for additional results. 
 }  
	\label{exp:compare_human}. 
 \vspace{-5mm}
\end{figure*}

\subsection{Efficient FVV Generation}\label{sec:compression}

In our methodology, we aim to produce compact FVV (Free-Viewpoint Video) content. This is achieved by encoding the representations, which have been previously trained, in a highly efficient manner. To facilitate this, we employ established, commercial video codecs known for their efficiency. The process involves a necessary transformation of our representations, ensuring their compatibility with the adopted video codecs.
%

%

We start by linearly normalizing the numbers so they fall between 0 and 1. According to our statistical analysis, in most scenarios, 99.5\% of density and feature values fall in the ranges $[-5,30]$ and $[-20,20]$, respectively. Any numbers that fall outside of these ranges are clipped to 0 or 1. Following this, we quantize the normalized values into 12-bit integers. 
We use the density activation function from DVGO paired with a threshold $\tau_\alpha$ to generate a grid mask and then cull empty space from density grids by setting the value in the mask to zero. This eliminates unnecessary temporal changes.
Since, feature planes have multiple channels, while compressed video has only a single channel, we flatten (or re-arrange) each channel of a feature plane onto a 2D single-channel image while preserving the 2D spatial continuity. 
Fig.~\ref{fig:compression} illustrates our compression pipeline.

In summary, we will have four images for each frame, representing density, xy-plane, xz-plane, and yz-plane, respectively, where each type forms an image sequence that we compress using a video codec. 
Rendering an FVV also requires the weights of the decoding MLPs for all frame groups. Therefore, we quantize those weights into 16 bits and store them directly without additional compression, given that the size of the MLPs is already relatively small. 
One benefit of leveraging off-the-shelf video codecs is that there are a lot of available options for software and hardware acceleration that can facilitate the efficient decoding of this kind of content.
In TeTriRF, we use the High Efficiency Video Coding (HEVC), also known as H.265, to compress feature images. 
We adjust Constant Rate Factor (CRF), which is a quality-control setting in video encoding that balances video quality and file size.


\begin{figure}[t]
	\begin{center}
		\includegraphics[width=1.0\linewidth]{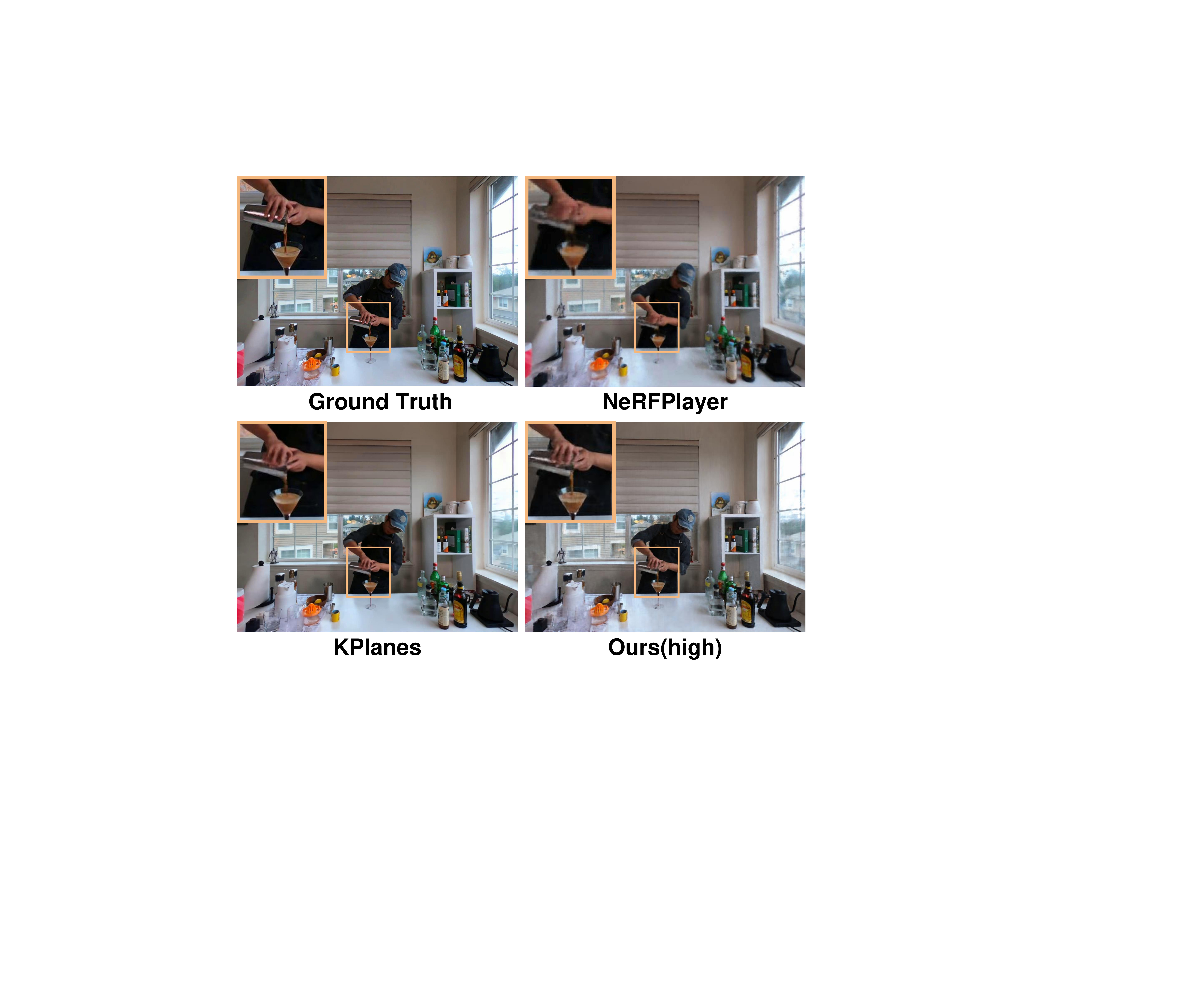} 
	\end{center}
 \vspace{-4mm}
	\caption{Qualitative results on the forward-facing scene in DyNeRF dataset. Visual comparisons on \textit{coffee martini}.} 
	\label{exp:compare_DyNeRF}
  \vspace{-4mm}
\end{figure}

\begin{table}[t]
\begin{tabular}{B|BBBBBB}\hline
                 & \multicolumn{6}{B}{DyNeRF Dataset}                                               \\ \midrule 
                 & PSNR    & SSIM  & LPIPS & T.T.  & R.T.  & Size  \\ \midrule
NeRFPlayer & 30.293 & 0.909 & 0.309 & 0.25  & 3.5 & 2427  \\
KPlanes   & 31.38 & 0.940 & 0.212 & 0.57  & 11.5 & 539   \\ \midrule
Ours(low)       & 28.71 & 0.867 & 0.321 & \multirow{2}{*}{0.65} & \multirow{2}{*}{0.24} & 21.46 \\
Ours(high)       & 30.43 & 0.906 & 0.248 &  &   & 62.5 \\ \hline
\bottomrule

\end{tabular}
\vspace{-1mm}
\caption{Comparison on the forward facing dataset DyNeRF. 
The training time (T.T. in minutes), rendering time (R.T. in seconds) and model size (Size) are averaged out over the number of frames.}
\label{tab: 2}
\vspace{-3mm}
\end{table}

\begin{figure*}[t]
	\begin{center}
		\includegraphics[width=1.0\linewidth]{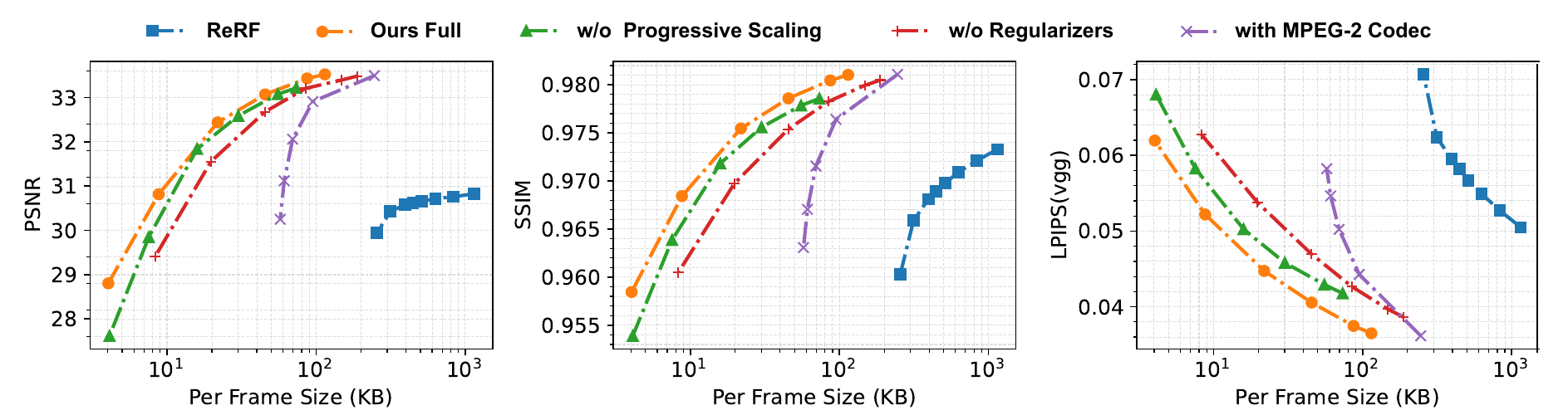} 
	\end{center}
 \vspace{-6mm}
	\caption{Rate-distortion curves. TeTriRF outperforms alternative versions and ReRF. Disabling progressive scaling (PS) or group-based regularization (Reg) reduces TeTriRF's performance. Even with MPEG-2, TeTriRF excels in compressing dynamic scenes. In the first two line graphs, the closer to the top left corner, the better; in the last one, the bottom left corner is optimal.}  
	\label{exp:ablation_all}
 \vspace{-4mm}
\end{figure*}
\begin{figure*}[t]
	\begin{center}
		\includegraphics[width=1.0\linewidth]{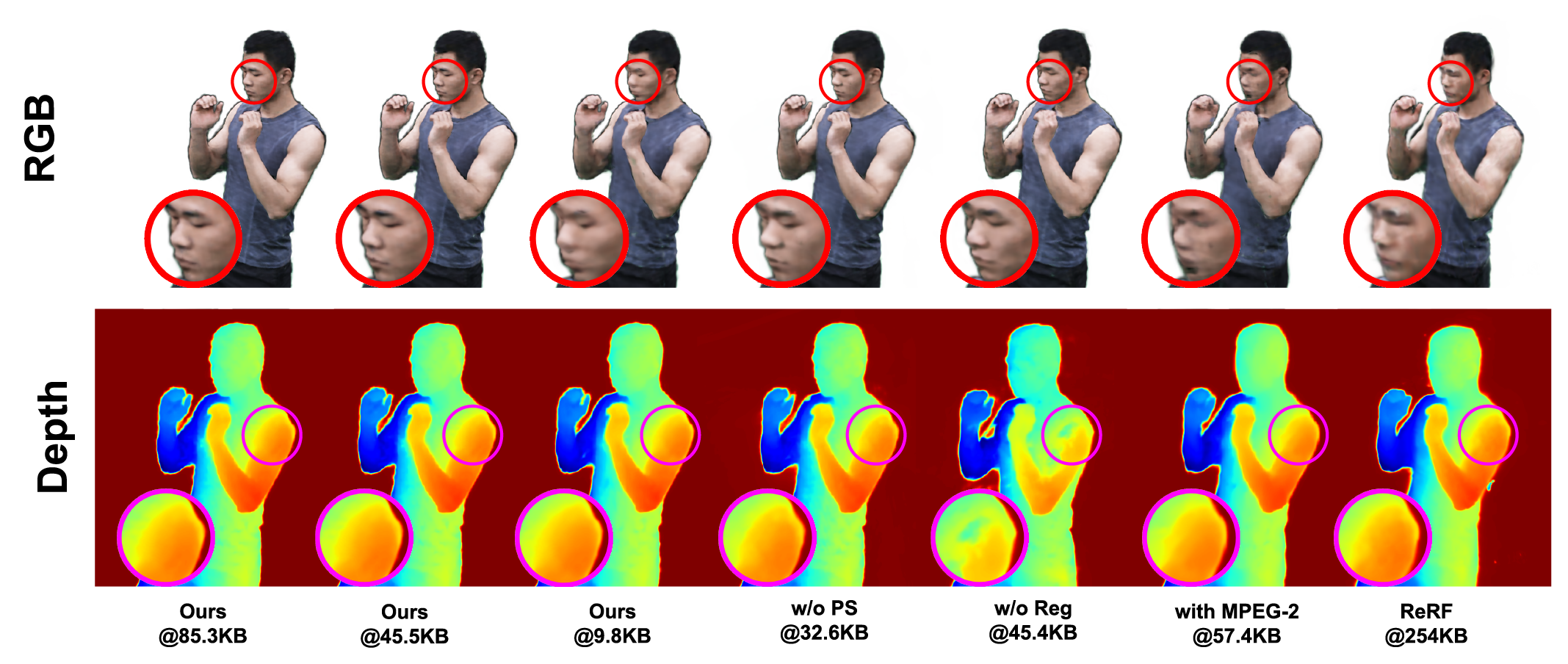} 
	\end{center}
  \vspace{-6mm}
	\caption{Qualitative results of complete TeTriRF model, its variants and ReRF. The variants are compared at approximately matched sizes.}  
	\label{exp:ablation_visual}
  \vspace{-2mm}
\end{figure*}

\section{Experiments}

We begin by comparing our method both quantitatively and qualitatively with previous works (Sec.~\ref{sec:exp_compare}). Subsequently, we present extensive ablation studies to validate the components we propose (Sec.~\ref{sec:exp_eval}). 
The default group size is $N=20$.
For more detailed information and additional results, please refer to our supplementary materials. 

In our experiments, we use three datasets: NHR~\cite{wu2020multi}, ReRF~\cite{wang2023rerf}, and DyNeRF~\cite{Li2021Neural3V}. The first two consist of human-centric dynamic scenes, while DyNeRF contains forward-facing dynamic scenes.

\subsection{Comparison} \label{sec:exp_compare}

%
For the human-centric scenes, our method is contrasted with four contemporary dynamic NeRF techniques: KPlanes \cite{fridovich2023k}, HumanRF \cite{isik2023humanrf}, TiNeuVox \cite{fang2022TiNeuVox}, and ReRF \cite{wang2023rerf}. For forward-facing scenes, our approach is compared with KPlanes \cite{fridovich2023k} and NeRFPlayer \cite{song2023nerfplayer}. 
For a fair comparison, we use their official codes and align the experimental setups for different datasets. Details can be found in the supplementary materials. 
%
We evaluate two versions of TeTriRF: one compressed using a high-quality option in the video codec (denoted as 'Ours(high)' with $CRF$=20), and another compressed with a lower quality option (denoted as 'Ours(low)' with $CRF$=33).

\textbf{Experiment Protocol.} 
For consistency and comparability, we limit the scope of training and evaluation to the initial 200 frames of each scene unless stated otherwise. We specify viewpoints 5 and 41 in the NHR dataset, viewpoints 6 and 39 in the ReRF dataset, and viewpoint 0 in the DyNeRF dataset as the test views, which are excluded during training. All models are benchmarked on an NVIDIA V100 GPU.


\textbf{Evaluation metrics.} 
Our evaluation framework focuses on three aspects: image quality, running time, and storage. For image quality, we use three standards: Peak Signal-to-Noise Ratio (PSNR), Structural Similarity Index (SSIM), and Learned Perceptual Image Patch Similarity (LPIPS) with VGG backbone. We assess storage by calculating the average model size per frame in kilobytes (KB), which is essential for rendering. We measure time efficiency by the average training time per frame in minutes (T.T.) and rendering time per frame in seconds (R.T.). Please note that we evaluate the rendering time in the Python implementation, which includes system overhead. Therefore, it only provides a relative comparison, not the actual rendering time in efficient implementations.

\textbf{Results.} 
In the assessment of the human-centric dataset detailed in Tab.~\ref{tab: 1}, `Ours(low)` demonstrates rendering quality that is on par with those of the KPlanes, TiNeuVox. Notably, it achieves this while requiring significantly less storage, over two orders of magnitude lower. 
Moreover, it offers a rendering time that is at least twice as fast as the most storage efficient SOTA method ReRF \cite{wang2023rerf}. `Ours(high)` yields further improvements: it not only enhances image quality and makes the method at least competitive with the SOTA methods but also surpasses the other methods in terms of running time and sizes. 

Tab.~\ref{tab: 2} shows the results on the DyNeRF dataset. In this table, we report the rendering quality of NeRFPlayer as it appears in the original paper, and additionally provide test results for running time and storage size, which were obtained by executing the official code.
`Ours(high)` delivers a rendering quality comparable to that of the other models while achieving significantly better time efficiency and requiring less storage space. 

Qualitative comparison between the baselines and our method can be found in Fig.~\ref{exp:compare_human} and Fig.~\ref{exp:compare_DyNeRF}. 
Our approach demonstrates the capability in handling intricate details of highly dynamic objects with a reduced model size. For instance, in Fig.~\ref{exp:compare_human}, our method can effectively capture details of a rapidly spinning basketball. Moreover, in Fig.~\ref{exp:compare_DyNeRF}, our method can also preserve more details accurately compared to the others, as evident in the clear geometry of fingers and the coffee flow.

\subsection{Evaluation}  \label{sec:exp_eval}

\textbf{Ablation Study} We evaluate the progressive scaling (PS) module and group training regularization (Reg) module by disabling them one at a time during training to analyze their individual contribution to the complete TeTriRF model.
For this ablation study, we selected the 'sport1' scene from the NHR dataset. 
In our experiments, we also replace the H265 codec with the MPEG-2 codec to assess how video codecs with different efficiency affect the compression rate and rendering quality. 
MPEG-2, being an earlier video encoding technology, has simpler algorithms that are akin to those used in ReRF's compression algorithm. By doing this, we aim to draw a comparison between our hybrid representation and the 3D voxel grid utilized in ReRF. 
Fig.~\ref{exp:ablation_all} illustrates the rate-distortion curves for various settings. Additionally, we have included the curve for ReRF~\cite{wang2023rerf} as a point of comparison. 
%
%
The exclusion of either PG or Reg leads to a deterioration in the performance of TeTriRF.
Even when employing a basic video codec like MPEG-2, TeTriRF still manages to outperform ReRF. This suggests that our proposed hybrid representation offers advantages in compressing dynamic scenes. 

Fig.~\ref{exp:ablation_visual} presents a qualitative comparison between the complete TeTriRF models at varying sizes and their variants in similar sizes.
In the absence of the PS module, the variant generates density floaters around the geometry surfaces, resulting in a marginally blurred RGB image. 
The variant lacking Reg struggles with geometry reconstruction, primarily due to insufficient temporal consistency. 
Both ReRF and the variant using MPEG-2 display significant inadequacies at this level of storage.

\begin{table}[]
\begin{tabular}{c|ccccc}
\hline
           & $\mathbf{P}_{xy}$  & $\mathbf{P}_{xz}$     & $\mathbf{P}_{yz}$     & $\mathbf{V}_{\sigma}$   & MLPs \\ \hline
Ours(low)  & 431  & 315.2  & 577.6  & 558.2 & \multirow{2}{*}{490} \\
Ours(high) & 4032 & 2758.6 & 5353.8 & 3350  &                      \\ \hline\hline
\end{tabular}
\vspace{-2mm}
\caption{Analysis of storage components in `Ours(low)' and `Ours(high)' based on 200 frames of sport1 scene (Values in KB).}
\label{tab:storage_breakdown}
\vspace{-2mm}
\end{table}

\textbf{Storage Breakdown.} We analyze the storage components of both `Ours(low)` and `Ours(high)`, breaking down each component within them. The data is compiled from statistics on 200 frames of the `sport1` scene.  Table~\ref{tab:storage_breakdown} shows the results. In the table, we exclude their metadata (e.g. the bounding box), which is less than 1KB..

\begin{figure}[t]
	\begin{center}
		\includegraphics[width=1.0\linewidth]{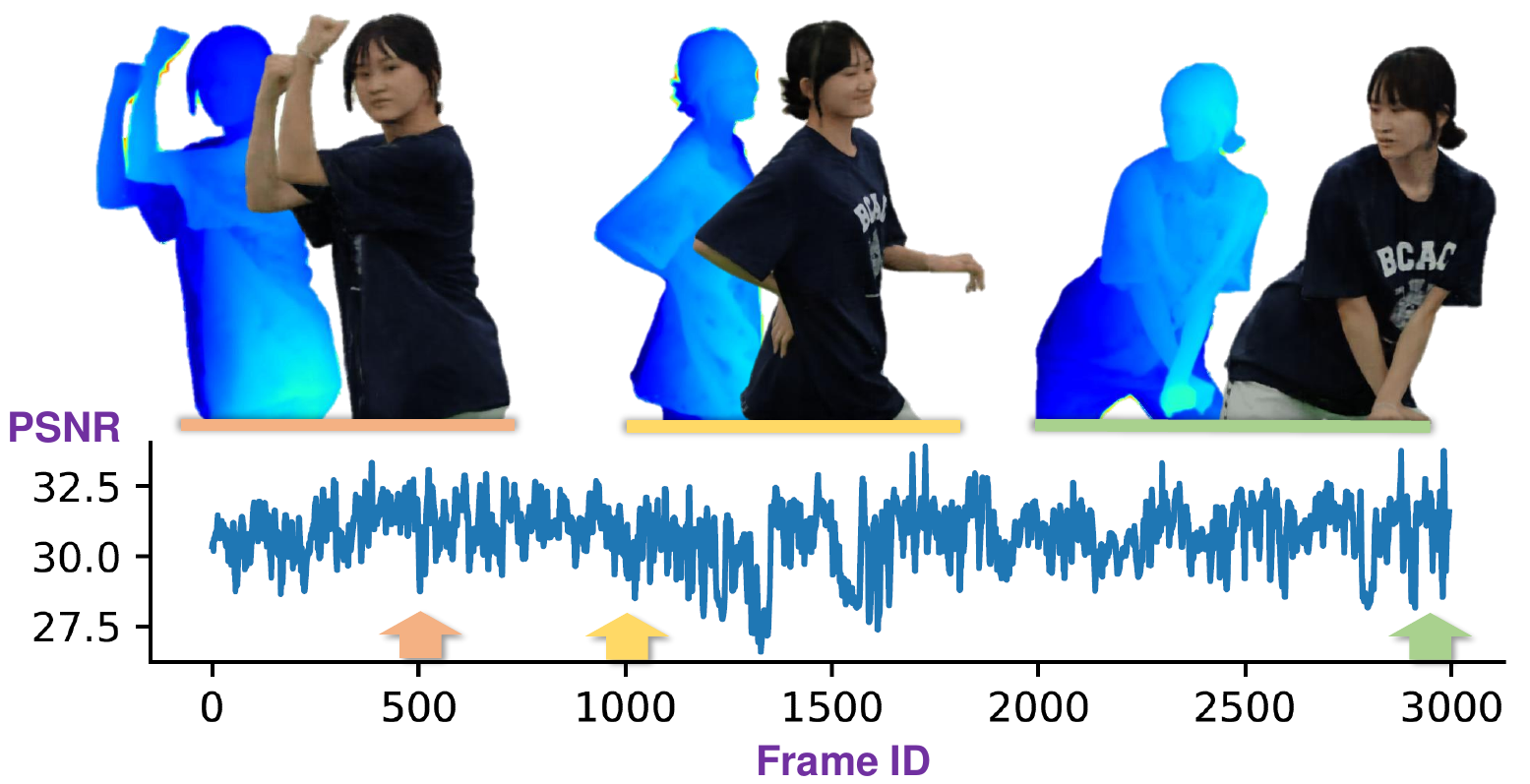} 
	\end{center}
    \vspace{-6mm}
	\caption{TeTriRF performance on the Kpop sequence, showcasing the first 3000 frames. 
 }  
 \vspace{-4mm}
	\label{exp:long_sequence}
\end{figure}
\textbf{Long Sequence.} 
We tested the TeTriRF on an ultra-long sequence, specifically the `Kpop` sequence in the ReRF dataset. Fig.~\ref{exp:long_sequence} shows the per-frame PSNR curve of TeTriRF over time under `ours(high)` setting. Despite the quality drop in the middle due to incorrect masks from the dataset, TeTriRF successfully resumes its correct trajectory without accumulating errors once the data returns to normal. 
\section{Conclusion}

The proposed innovative training scheme elevates the sequential representation's temporal coherence and low-entropy characteristics, resulting in a dramatic enhancement of compression efficiency. 
Our evaluation has demonstrated  the compactness and effectiveness  of TeTriRF's hybrid representation that plays an important part in compression process.
Leveraging our compression pipeline, TeTriRF is able to support long-duration FVV experiences, while remarkably minimizing storage requirements. 
TeTriRF's rendering pipeline is efficient, straightforward, and opens the door to leveraging GLSL shaders for our method, paving the way for real-time performance on diverse devices, supported by hardware-accelerated video decoding.
%
Training efficiency, fast rendering speed, and compact data storage of TeTriRF enable photo-realistic FVV applications in AR/VR contexts.

\textbf{Acknowledgments.} {\raggedleft This work was supported by the Flanders AI Research program. }

{
    \small
    \bibliographystyle{ieeenat_fullname}
    \bibliography{main}
}

\clearpage
\setcounter{page}{1}
\maketitlesupplementary

\section{Implementation Details}
\textbf{Model Configuration.} 
In our setup, all frames within a sequence utilize a common bounding box that defines their world space. These bounding boxes are derived based on the camera configurations. 
For object-centric datasets (NHR and ReRF), the world size is set to $120^3$, while for the DyNeRF dataset, it is $210^3$. 
We determine the feature plane resolution as three times the world size. Specifically, this results in approximately $360\times360$ for the NHR and ReRF datasets and $600\times600$ for the DyNeRF dataset, aiming to capture high-frequency signals effectively. 
Each feature plane comprises $h=10$ channels, leading to a concatenated feature vector for each 3D point with a dimensionality of $30$. 
The viewing directions undergo positional encoding with 4 frequency levels. We combine these encoded viewing directions with point feature vectors to serve as inputs for the MLP decoder $\Phi$. 
The decoder $\Phi$ is a three-layer multilayer perceptron, having a width of 128. It outputs the RGB value for the sampled point. 
A unique MLP decoder is allocated to each frame group, facilitating shared learning across the frames in a group.

\textbf{Training.}
For training, we employ the Adam optimizer~\cite{kingma2014adam} to update the density grids, tri-planes, and MLP decoder weights. 
The respective learning rates for these components are set to $1.5e^{-1}$ for the density grids and tri-planes, and $1e^{-3}$ for the MLP decoder. 
We implement group-based regularization with weights $\lambda_1=1e^{-3}$ and $\lambda_2=2e^{-3}$. 
Each training batch processes 17800 sampled rays from the dataset, and we conduct 40000 training iterations for each group. 
In our progressive scaling approach, the hybrid representation is upscaled by a factor of two at specific iterations: $[1000, 2000, 3000, 4000]$ during the first pass, and $[9000, 11000, 13000]$ during the second. We downscale the resolutions at the first and $7000$-th iterations. Full resolutions are achieved at the $4000$-th and $13000$-th iterations. 
Every 1000 iterations, we update the occupancy grids $V_{o}$ based on the density grids $V_{\sigma}$, fomulated as
\begin{equation}
\begin{split}
V_{o} = \rho(\kappa(1-\frac{1}{1+\exp(V_{\sigma})}), \lambda_{th}).
\end{split}
\label{eq:fetch2}
\end{equation}
Here, $\kappa(\cdot)$ represents a 3D max pooling function with a $3\times3$ kernel, and $\rho(\cdot)$ is a thresholding function that outputs 1 if the grid element is greater than $\lambda_{th}=1e^{-4}$, otherwise 0, indicating occupancy. 
At the $13000$-th iteration, we filter out rays that do not intersect with any objects according to the current occupancy grid.

\begin{figure*}[t]
	\begin{center}
		\includegraphics[width=1.0\linewidth]{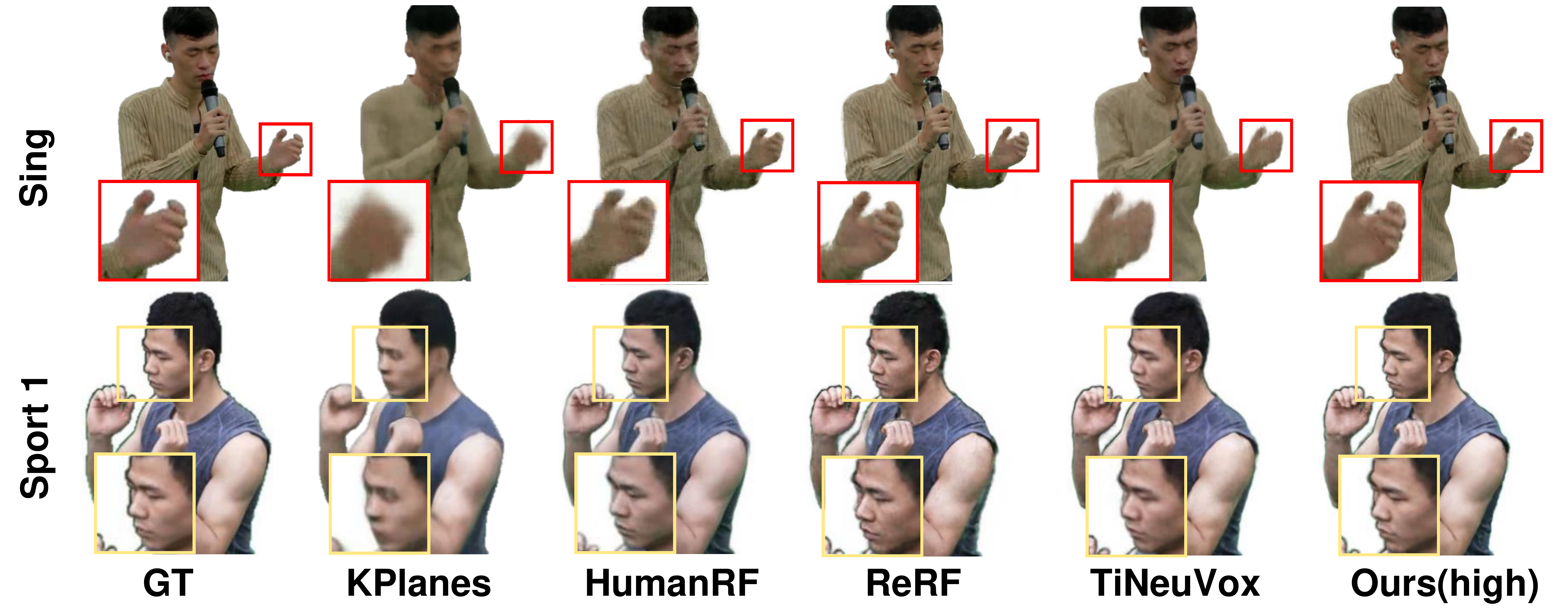} 
	\end{center}
    \vspace{-6mm}
	\caption{Extra qualitative results on `Sing` from ReRF dataset~\cite{wang2023rerf} and `Sport1` from NHR dataset~\cite{wu2020multi}.}  
	\label{exp:comp_human}
\end{figure*}

\begin{table}[]
\begin{tabular}{lc|cccc}
\hline
                                                 & \multicolumn{1}{l|}{} & PSNR  & SSIM  & LPIPS & \makecell{Size \\ (KB)} \\ \hline
\multicolumn{1}{l|}{\multirow{6}{*}{\rotatebox{90}{sport1}}}     & K-plane               & 30.40 & 0.962 & 0.0615     & 2986           \\
\multicolumn{1}{l|}{}                            & HumanRF               & 32.39 & 0.885 & 0.0318     & 2852           \\
\multicolumn{1}{l|}{}                            & TiNeuVox              & 30.54 & 0.961 & 0.0831     & 5580           \\
\multicolumn{1}{l|}{}                            & ReRF                  & 30.83 & 0.973 & 0.0505     & 1113           \\ \cline{2-6} 
\multicolumn{1}{l|}{}                            & Ours(low)             & 31.79 & 0.969 & 0.0516     & 11.01          \\
\multicolumn{1}{l|}{}                            & Ours(high)            & 33.41 & 0.980 & 0.0389     & 79.92          \\ \hline
\multicolumn{1}{l|}{\multirow{6}{*}{\rotatebox{90}{sport2}}}     & K-plane               & 32.10 & 0.975 & 0.0472     & 2986           \\
\multicolumn{1}{l|}{}                            & HumanRF               & 33.04 & 0.889 & 0.0316     & 2852           \\
\multicolumn{1}{l|}{}                            & TiNeuVox              & 32.97 & 0.972 & 0.0568     & 5580           \\
\multicolumn{1}{l|}{}                            & ReRF                  & 31.83 & 0.976 & 0.0487     & 1316           \\ \cline{2-6} 
\multicolumn{1}{l|}{}                            & Ours(low)             & 31.75 & 0.973 & 0.0498     & 10.56          \\
\multicolumn{1}{l|}{}                            & Ours(high)            & 34.14 & 0.983 & 0.0383     & 75.85          \\ \hline
\multicolumn{1}{l|}{\multirow{6}{*}{\rotatebox{90}{sport3}}}     & K-plane               & 30.20 & 0.962 & 0.0610     & 2986           \\
\multicolumn{1}{l|}{}                            & HumanRF               & 32.11 & 0.885 & 0.0328     & 2852           \\
\multicolumn{1}{l|}{}                            & TiNeuVox              & 30.11 & 0.960 & 0.0696     & 5580           \\
\multicolumn{1}{l|}{}                            & ReRF                  & 30.89 & 0.976 & 0.0473     & 1243           \\ \cline{2-6} 
\multicolumn{1}{l|}{}                            & Ours(low)             & 30.38 & 0.967 & 0.0546     & 12.96          \\
\multicolumn{1}{l|}{}                            & Ours(high)            & 32.90 & 0.980 & 0.0394     & 94.58          \\ \hline
\multicolumn{1}{l|}{\multirow{6}{*}{\rotatebox{90}{basketball}}} & K-plane               & 28.02 & 0.957 & 0.0822     & 2986           \\
\multicolumn{1}{l|}{}                            & HumanRF               & 30.09 & 0.829 & 0.0469     & 2852           \\
\multicolumn{1}{l|}{}                            & TiNeuVox              & 28.18 & 0.956 & 0.0991     & 5580           \\
\multicolumn{1}{l|}{}                            & ReRF                  & 27.82 & 0.963 & 0.0747     & 1208           \\ \cline{2-6} 
\multicolumn{1}{l|}{}                            & Ours(low)             & 27.79 & 0.957 & 0.0806     & 12.53          \\
\multicolumn{1}{l|}{}                            & Ours(high)            & 29.85 & 0.970 & 0.0649     & 90.97          \\ \hline \hline
\end{tabular}
\caption{Per-scene results on NHR dataset~\cite{wu2020multi}. Values are averaged out over the number of frames in each scene.}
\label{tab:perscene_NHR}
\end{table}


\begin{table}[]
\begin{tabular}{lc|cccc}
\hline
                                           & \multicolumn{1}{l|}{} & PSNR  & SSIM  & LPIPS & \makecell{Size \\ (KB)} \\ \hline
\multicolumn{1}{l|}{\multirow{6}{*}{\rotatebox{90}{box}}}  & K-plane               & 27.96 & 0.952 & 0.0836     & 2986           \\
\multicolumn{1}{l|}{}                      & HumanRF               & 29.07 & 0.884 & 0.0614     & 2852           \\
\multicolumn{1}{l|}{}                      & TiNeuVox              & 31.11 & 0.962 & 0.0633     & 5580           \\
\multicolumn{1}{l|}{}                      & ReRF                  & 30.97 & 0.972 & 0.0516     & 925            \\ \cline{2-6} 
\multicolumn{1}{l|}{}                      & Ours(low)             & 27.94 & 0.955 & 0.0655     & 11.86          \\
\multicolumn{1}{l|}{}                      & Ours(high)            & 31.39 & 0.968 & 0.0498     & 70.01          \\ \hline
\multicolumn{1}{l|}{\multirow{6}{*}{\rotatebox{90}{kpop}}} & K-plane               & 26.95 & 0.954 & 0.0984     & 2986           \\
\multicolumn{1}{l|}{}                      & HumanRF               & 28.84 & 0.901 & 0.0682     & 2852           \\
\multicolumn{1}{l|}{}                      & TiNeuVox              & 27.22 & 0.952 & 0.0887     & 5580           \\
\multicolumn{1}{l|}{}                      & ReRF                  & 31.94 & 0.976 & 0.0436     & 725            \\ \cline{2-6} 
\multicolumn{1}{l|}{}                      & Ours(low)             & 27.03 & 0.964 & 0.0678     & 13.11          \\
\multicolumn{1}{l|}{}                      & Ours(high)            & 30.25 & 0.977 & 0.0526     & 80.1           \\ \hline
\multicolumn{1}{l|}{\multirow{6}{*}{\rotatebox{90}{sing}}} & K-plane               & 28.52 & 0.931 & 0.1009     & 2986           \\
\multicolumn{1}{l|}{}                      & HumanRF               & 27.84 & 0.846 & 0.0874     & 2852           \\
\multicolumn{1}{l|}{}                      & TiNeuVox              & 28.28 & 0.929 & 0.0956     & 5580           \\
\multicolumn{1}{l|}{}                      & ReRF                  & 28.11 & 0.937 & 0.0688     & 879            \\ \cline{2-6} 
\multicolumn{1}{l|}{}                      & Ours(low)             & 27.84 & 0.931 & 0.0818     & 10.19          \\
\multicolumn{1}{l|}{}                      & Ours(high)            & 28.91 & 0.942 & 0.0669     & 64.92          \\ \hline\hline
\end{tabular}

\caption{Per-scene results on ReRF dataset~\cite{wang2023neural}. Values are averaged out over the number of frames in each scene.}
\label{tab:perscene_ReRF}
\end{table}

\begin{table}[]
\begin{tabular}{cc|cccc} \hline
\multicolumn{1}{l}{}                                   & \multicolumn{1}{l|}{} & PSNR  & SSIM  & LPIPS & \makecell{Size \\ (KB)} \\ \hline
\multicolumn{1}{c|}{\multirow{4}{*}{\rotatebox{90}{\makecell{flame\\samon}}}}      & K-plane               & 30.57 & 0.925 & 0.2105     & 539            \\
\multicolumn{1}{c|}{}                                  & NeRFPlayer            & 26.14 & 0.849 & 0.3790     & 2427           \\ \cline{2-6} 
\multicolumn{1}{c|}{}                                  & Ours(low)             & 26.69 & 0.830 & 0.3486     & 26.70          \\
\multicolumn{1}{c|}{}                                  & Ours(high)            & 28.05 & 0.872 & 0.2727     & 73.78          \\ \hline
\multicolumn{1}{c|}{\multirow{4}{*}{\rotatebox{90}{\makecell{flame\\steak}}}}      & K-plane               & 32.88 & 0.957 & 0.2021     & 539            \\
\multicolumn{1}{c|}{}                                  & NeRFPlayer            & 27.36 & 0.867 & 0.3550     & 2427           \\ \cline{2-6} 
\multicolumn{1}{c|}{}                                  & Ours(low)             & 30.11 & 0.891 & 0.3031     & 18.90          \\
\multicolumn{1}{c|}{}                                  & Ours(high)            & 32.13 & 0.929 & 0.2295     & 56.12          \\ \hline
\multicolumn{1}{c|}{\multirow{4}{*}{\rotatebox{90}{\makecell{coffee\\martini}}}}   & K-plane               & 30.22 & 0.925 & 0.2113     & 539            \\
\multicolumn{1}{c|}{}                                  & NeRFPlayer            & 32.05 & 0.938 & 0.2790     & 2427           \\ \cline{2-6} 
\multicolumn{1}{c|}{}                                  & Ours(low)             & 26.28 & 0.822 & 0.3626     & 26.80          \\
\multicolumn{1}{c|}{}                                  & Ours(high)            & 27.26 & 0.865 & 0.2890     & 76.37          \\ \hline
\multicolumn{1}{c|}{\multirow{4}{*}{\rotatebox{90}{\makecell{cut roasted \\beef}}}} & K-plane               & 32.08 & 0.943 & 0.2196     & 539            \\
\multicolumn{1}{c|}{}                                  & NeRFPlayer            & 31.83 & 0.928 & 0.2870     & 2427           \\ \cline{2-6} 
\multicolumn{1}{c|}{}                                  & Ours(low)             & 29.60 & 0.887 & 0.3035     & 18.54          \\
\multicolumn{1}{c|}{}                                  & Ours(high)            & 31.57 & 0.923 & 0.2374     & 56.13          \\ \hline
\multicolumn{1}{c|}{\multirow{4}{*}{\rotatebox{90}{\makecell{cook\\spinach}}}}     & K-plane               & 30.87 & 0.938 & 0.2212     & 539            \\
\multicolumn{1}{c|}{}                                  & NeRFPlayer            & 32.06 & 0.930 & 0.2840     & 2427           \\ \cline{2-6} 
\multicolumn{1}{c|}{}                                  & Ours(low)             & 29.40 & 0.882 & 0.3079     & 20.01          \\
\multicolumn{1}{c|}{}                                  & Ours(high)            & 31.41 & 0.919 & 0.2398     & 60.00          \\ \hline
\multicolumn{1}{c|}{\multirow{4}{*}{\rotatebox{90}{\makecell{sear\\steak}}}}       & K-plane               & 31.69 & 0.955 & 0.2057     & 539            \\
\multicolumn{1}{c|}{}                                  & NeRFPlayer            & 32.31 & 0.940 & 0.2720     & 2427           \\ \cline{2-6} 
\multicolumn{1}{c|}{}                                  & Ours(low)             & 30.19 & 0.892 & 0.2994     & 17.81          \\
\multicolumn{1}{c|}{}                                  & Ours(high)            & 32.18 & 0.931 & 0.2245     & 52.60          \\ \hline\hline
\end{tabular}
\caption{Per-scene results on DyNeRF dataset~\cite{Li2021Neural3V}. Values are averaged out over the number of frames in each scene.}
\label{tab:perscene_DyNeRF}
\end{table}

\begin{figure*}[t]
	\begin{center}
		\includegraphics[width=1.0\linewidth]{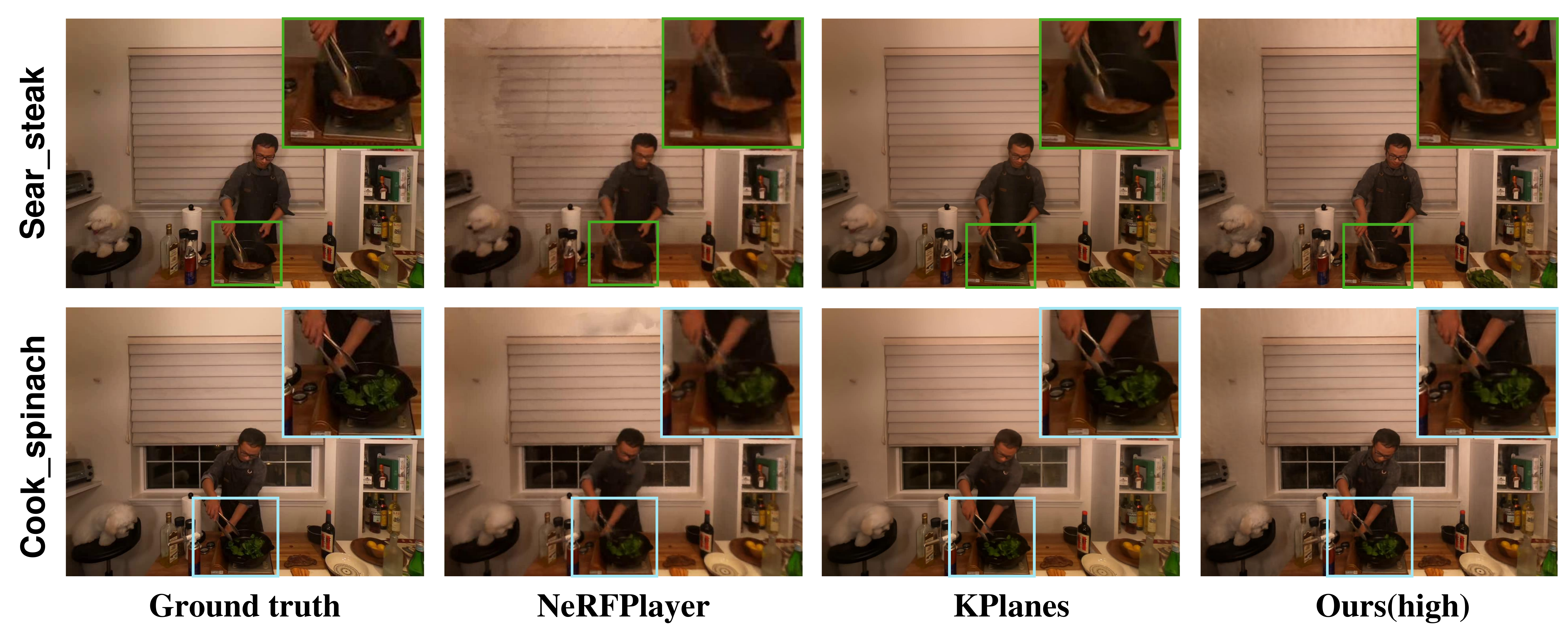} 
	\end{center}
    \vspace{-6mm}
	\caption{Extra qualitative results on DyNeRF dataset~\cite{Li2021Neural3V}.}  
	\label{exp:comp_n3d}
\end{figure*}

\section{Comparison Setups}
We compared TeTriRF with several contemporary dynamic NeRF techniques, including KPlanes \cite{fridovich2023k}, HumanRF \cite{isik2023humanrf}, TiNeuVox \cite{fang2022TiNeuVox}, and ReRF~\cite{wang2023rerf}. For forward-facing scenes, TeTriRF was additionally compared with NeRFPlayer~\cite{song2023nerfplayer}.

\textbf{KPlanes.} 
We use the official implementation from NeRFStudio~\cite{tancik2023nerfstudio}. The KPlanes model was jointly trained on the entire sequence for 50,000 iterations, using a grid size of $256^3$ and a time resolution of $100$, as recommended.

\textbf{HumanRF.} 
We employed their official code for our experiments. Two hundred frames were trained jointly over 50,000 iterations. Initially, occupancy grids were generated using foreground masks as outlined in \cite{isik2023humanrf}, followed by their prescribed training steps.

\textbf{TiNeuVox.} 
We use their official code. Due to memory constraints, each sequence was split into eight groups of 25 frames each. We used a grid size of $180^3$ and trained each group for 30,000 iterations.

\textbf{ReRF.} 
The official code and default settings were used in our experiments, compressing sequences with a quality factor of $99$.

\textbf{NeRFPlayer.} 
We relied on the quantitative results reported in the original paper~\cite{song2023nerfplayer} and conducted qualitative analyses using the official NeRFStudio implementation based on Nerfacto under default settings.

\textbf{Ours.} 
Following the configurations detailed in our implementation section, we leverage the FFMPEG software with the libx265 codec for compressing the feature and density image sequences.

\section{More Results}
Table~\ref{tab:perscene_NHR}, Table~\ref{tab:perscene_ReRF}, and Table~\ref{tab:perscene_DyNeRF} provide the detailed results for each scene. Figure~\ref{exp:comp_human} and Figure~\ref{exp:comp_n3d} demonstrate the qualitative comparison on three datasets.


\end{document}